# Mutual-energy inner product optimization method for constructing feature coordinates and image classification in Machine Learning

Yuanxiu Wang
iamwangyuanxiu@berkeley.edu
Department of EECS, University of California, Berkeley, CA, 94720, USA

**Abstract:**
As a key task in machine learning, data classification is essentially to find a suitable coordinate system to represent data features of different classes of samples. This paper proposes the mutual-energy inner product optimization method for constructing a feature coordinate system. First, by analyzing the solution space and eigenfunctions of partial differential equations describing a non-uniform membrane, the mutual-energy inner product is defined. Second, by expressing the mutual-energy inner product as a series of eigenfunctions, it shows a significant advantage of enhancing low-frequency features and suppressing high-frequency noise, compared with the Euclidean inner product. And then, a mutual-energy inner product optimization model is built to extract data features, and convexity and concavity properties of its objective function are discussed. Next, by combining the finite element method, a stable and efficient sequential linearization algorithm is constructed to solve the optimization model. This algorithm only solves equations including positive definite symmetric matrix and linear programming with a few constraints, and its vectorized implementation is discussed. Finally, the mutual-energy inner product optimization method is used to construct feature coordinates, and multi-class Gaussian classifiers are trained on the MINST training set. Good prediction results of Gaussian classifiers are achieved on the MINST test set.

## 1. Introduction

In machine learning, data classification plays a very important role. Up to now, a large number of data classification methods have emerged and powered the development of machine learning and its practical applications in different domains [1], such as image detection [2], speech recognition [3], text understanding [4], disease diagnosis [5,6], and financial prediction [7].

Currently, popular data classification methods include Support Vector Machine (SVM) [8,9], Decision Tree (DT) [10,11], Naive Bayes (NB) [12], K-Nearest Neighbors (KNN) [13], Random Forest (RF) [14], Deep Learning (DL) [15] and Deep Reinforcement Learning (DRL) [16]. SVM is based on optimization theory [17]. DL is implemented through a multilayer neural network under the guidance of optimization techniques, such as the stochastic gradient descent algorithm [18]. DRL combines DL with Reinforcement Learning, and it is effective in the real-time scenarios [16]. The others fall in the title of statistical methods [19, 20].

Many comparative studies are employed to evaluate these classification methods by analyzing their accuracies, time costs, stability, sensitivity, as well as advantages and disadvantages [21-23]. SVM is efficient when there is a clear margin of separation between classes, but the choice of its kernel function is difficult, and it does not work with noisy data sets [23, 24]. DL is developing rapidly, but its training is a very time-consuming process because a large number of parameters need to be optimized through the stochastic gradient descent algorithm. Besides, some hyperparameters in DL are set empirically, such as the number of layers in the neural network, the number of nodes in each layer, and the learning rate, resulting in high sensitivity in performance dependent on hyperparameters and specific problems [25, 26]. KNN and DT are easy to apply. However, KNN requires calculation of the Euclidean distance between all points, leading to high computation cost. DT is unsuitable for continuous variables, and it has a problem of overfitting [23]. Other classical methods also obtained great successes [27, 28]. In order to improve classification accuracy, an ensemble learning scheme, such as AdaBoost [29, 30], Bagging [31], Stacking [32] and Gradient Boosting [33], is usually adopted to solve an intricate or large-scale problem [34, 35].

Inspired by the ability of our brain to recognize musical notes played by any musical instrument in a noisy environment, this paper proposes an optimization method for constructing feature coordinates for data classification by simulating a non-uniform membrane structure model. No matter how complex a musical instrument's structure is, or how different its vibration patterns are, when we listen to a piece of music played by an instrument, our brain can extract the fundamental tone of its vibration at every moment, and can recognize the beautiful melody as time goes by. Mathematically, this can be clearly explained. The vibration of the musical instrument at every moment is adaptively expanded on its own eigenfunction system, and our brain can grasp the lowest eigenvalue and its eigenfunction components corresponding to musical notes every moment, and enjoy the beautiful melody over time. In order to extract data features from complex samples, we simulate the adaptively generating process of an eigenfunction coordinate system of the musical instrument and build the mapping from data features to the low-frequency subspace of the eigenfunction system. Through analyzing the solution space and eigenfunctions of partial differential equations describing the vibration of a non-uniform membrane, which is a simple musical instrument, the mutual-energy inner product is defined and is used to extract data features. The introduction of the mutual-energy inner product not only can avoid generating eigenfunction system to reduce the computational complexity, but also can enhance feature information and filter out data noise, furthermore, can benefit the simplification of the data classifier training.

The full paper is divided into 6 sections. The first section briefly introduces popular data classification methods and research background. The second section analyzes the solution space of partial differential equations describing a non-uniform membrane, and defines the concept of the mutual-energy inner product. In the third section, by making use of the eigenvalues and eigenfunctions of the non-uniform membrane vibration equations, the mutual-energy inner product is expressed as a series of eigenfunctions, and its potential in data classification is pointed out for enhancing feature information and filtering out data noise. The fourth builds a mutual-energy inner product optimization model and discusses convexity and concavity properties of its objective function. The fifth designs a sequential linearization algorithm to solve the optimization model by combing Finite Element Method (FEM). In the sixth section, the mutual-energy inner product optimization method for constructing feature coordinates is applied to a 2-D image classification problem, and numerical examples are given in combination with Gaussian classifiers and the handwritten digit MINST dataset. Finally, we summarize the full paper and introduce future scope of the work.

## 2. Mutual-Energy Inner Product

Consider the linear partial differential equations

$$\begin{cases} L[u(\vec{x})] = f(\vec{x}) & \vec{x} \in \Omega \\ l[u(\vec{x})] = 0 & \vec{x} \in \Gamma \end{cases} \tag{1}$$

Where, $L[u(\vec{x})]$ is a homogeneous linear self-adjoint differential operator; $f(\vec{x})$ is a piecewise continuous function; $\Omega$ is the domain of definition, with a boundary $\Gamma$; and $l[u(\vec{x})]$ is a homogeneous linear differential operator on the boundary $\Gamma$, describing the Robin boundary condition.

$$\begin{cases} L[u(\vec{x})] = -\sum_{i=1}^{n} \frac{\partial}{\partial x_i}\left(p(\vec{x})\frac{\partial u(\vec{x})}{\partial x_i}\right) + q(\vec{x})u(\vec{x}) \\ l[u(\vec{x})] = p(\vec{x})\frac{\partial u(\vec{x})}{\partial n} + \sigma(\vec{x})u(\vec{x}) \end{cases} \tag{2}$$

Expression (2) can be regarded as static equilibrium equations of a simple elastic structure, such as a 1-D string, a 2-D membrane, and can be expanded to an n-dimensional problem. For a 2-dimensional problem, $\Omega$ is a domain occupied by a membrane with its boundary $\Gamma$; $p(\vec{x})$ and $q(\vec{x})$ stand for the elastic modulus and distributed support elastic coefficient of the membrane, respectively; $\sigma(\vec{x})$ is the support elastic coefficient on the boundary; $f(\vec{x})$ is an external force acting on the membrane; $u(\vec{x})$ is the deformations of the membrane due to $f(\vec{x})$, and has a piecewise continuous first-order derivative; $\partial u / \partial n$ is the derivative of the deformations in the outward-pointing normal direction of $\Gamma$. In this research, it is required that, $p(\vec{x})$, $q(\vec{x})$, $\sigma(\vec{x})$ are piecewise continuous functions, and $p(\vec{x}) > 0$, $q(\vec{x}) > 0$, $\sigma(\vec{x}) \geq 0$.

A structure subjected to an external force $f(\vec{x})$ will generate the deformation $u(\vec{x})$, and its deformation energy $E[u(\vec{x})]$ can be expressed as

$$E[u(\vec{x})] = \frac{1}{2}\int_{\Omega} f(\vec{x})u(\vec{x})d\Omega \tag{3}$$

If the structure is simultaneously subjected to another external force $g(\vec{x})$, then it will generate an additional deformation $v(\vec{x})$. The total deformation $u(\vec{x}) + v(\vec{x})$ satisfies the superposition principle due to the linearity of Expression (1). The deformation $v(\vec{x})$ can cause an additional work done by $f(\vec{x})$. Generally, the additional deformation energy $U[u(\vec{x}), v(\vec{x})]$ is called the mutual energy between $u(\vec{x})$ and $v(\vec{x})$ or mutual work between $f(\vec{x})$ and $g(\vec{x})$. The mutual energy describes the correlation of the two external forces, and can be expressed as

$$U[u(\vec{x}), v(\vec{x})] = \int_\Omega f(\vec{x}) v(\vec{x}) d\Omega \tag{4}$$

Substituting Expression (1) into Expression (4), by integrating by parts, we obtain

$$U[u(\vec{x}), v(\vec{x})] = \int_\Omega \left( \sum_{i=1}^n \frac{\partial u(\vec{x})}{\partial x_i} p(\vec{x}) \frac{\partial v(\vec{x})}{\partial x_i} + u(\vec{x}) q(\vec{x}) v(\vec{x}) \right) d\Omega + \int_\Gamma u(\vec{x}) \sigma(\vec{x}) v(\vec{x}) d\Gamma \tag{5}$$

Expression (5) is a bilinear functional. Comparing the Expressions (3, 4), we have

$$E[u(\vec{x})] = \frac{1}{2} U[u(\vec{x}), u(\vec{x})] \tag{6}$$

Due to $p(\vec{x}) > 0$, $q(\vec{x}) > 0$, $\sigma(\vec{x}) \geq 0$, according to the Expressions (5, 6), the mutual energy satisfies

$$U[u(\vec{x}), u(\vec{x})] \geq 0 \quad \left( if\ only\ if\quad u(\vec{x}) = 0,\ U[u(\vec{x}), u(\vec{x})] = 0 \right) \tag{7}$$

Expression (7) describes a simple physical phenomenon: when the elastic modulus of structural material is positive, if the structure deforms, deformation energy generates, otherwise, the deformation energy is zero.

Expression (5) also shows that the mutual energy is symmetrical and satisfies the commutative law. Combined with Expression (7), it can be inferred that the mutual energy satisfies the Cauchy-Schwarz inequality

$$\left( U[u(\vec{x}), v(\vec{x})] \right)^2 \leq U[u(\vec{x}), u(\vec{x})] \cdot U[v(\vec{x}), v(\vec{x})] \tag{8}$$

The Expressions (7, 8) show that the mutual energy can be regarded as an inner product of structural deformation functions. For simplicity, we use $\langle u, v \rangle_U$ and $\langle u, v \rangle$ to represent the mutual-energy inner product and the Euclidean inner product, respectively, that is,

$$\begin{cases} \langle u, v \rangle_U = U[u(\vec{x}), v(\vec{x})] \\ \langle u, v \rangle = \int_\Omega u(\vec{x}) v(\vec{x}) d\Omega \end{cases} \tag{9}$$

We define $\|u\|_2$ as the norm derived from $\langle u, v \rangle$, and $\|u\|_U$ as the norm derived from $\langle u, v \rangle_U$. Based on Expression (6), $\|u\|_U$ satisfies

$$\|u\|_U = \sqrt{2E[u(\vec{x})]} \tag{10}$$

$\|u\|_U$ is proportional to the square root of the deformation energy, and is also the energy norm. According to the Cauchy-Schwarz inequality (8), $\|u\|_U$ satisfies the triangle inequality

$$\|u + v\|_U \leq \|u\|_U + \|v\|_U \tag{11}$$

Based on Expression (1), when a structure is subjected to a piecewise continuous external force, its deformation function has piecewise continuous first-order derivatives on the domain $\Omega$ and satisfies the boundary condition $l[u(\vec{x})] = 0$. The set of these deformation functions can span a space $V(\Omega)$, which can be equipped either with the Euclidean inner product $\langle u, v \rangle$ or with the mutual-energy inner product $\langle u, v \rangle_U$.

In addition, applying the variational principle, Expression (1) can also be rewritten as the minimum energy principle expression

$$\min_{u(\vec{x})} \pi[u] = \frac{1}{2} \langle u, u \rangle_U - \langle f, u \rangle \tag{12}$$

Here, the feasible domain of $u(\vec{x})$ has piecewise continuous first-order derivatives on $\Omega$, and does not need to satisfy homogeneous boundary conditions.

## 3. Signal Processing Property of Mutual-Energy Inner Product

The eigenequation of $L[u(\vec{x})]$ can be written as

$$\begin{cases} L[u(\vec{x})] = \lambda u(\vec{x}) & \vec{x} \in \Omega \\ p(\vec{x})\dfrac{\partial u(\vec{x})}{\partial n} + \sigma(\vec{x})u(\vec{x}) = 0 & \vec{x} \in \Gamma \end{cases} \quad (13)$$

For Expression (13), its non-zero solutions $\varphi(\vec{x})$ and the corresponding coefficients $\lambda$ are called eigenfunctions and eigenvalues, respectively. These eigenfunctions and eigenvalues have the following properties due to $p(\vec{x}) > 0$, $q(\vec{x}) > 0$, $\sigma(\vec{x}) \geq 0$ [36].

(1) Expression (13) has infinite eigenvalues $\lambda_n$ and eigenfunctions $\varphi_n(\vec{x})$, i.e. $n = 1, 2 \cdots \infty$. If all eigenvalues are ranked like $\lambda_1 \leq \lambda_2 \cdots \lambda_n \leq \cdots$ then they satisfy $\lambda_1 > 0$ and $\lim_{n \to \infty} \lambda_n = +\infty$. Meanwhile, $\lambda_n$ has continuous dependence on $p(\vec{x})$, $q(\vec{x})$ and $\sigma(\vec{x})$, and will increase with the increase in $p(\vec{x})$, $q(\vec{x})$ and $\sigma(\vec{x})$.

(2) Normalized eigenfunctions $\varphi_n(\vec{x})$ satisfy the orthogonality condition (14), and can form a set of orthogonal and complete basis functions to span the deformation function space $V(\Omega)$.

$$\langle \varphi_m, \varphi_n \rangle = \begin{cases} 1 & m = n \\ 0 & m \neq n \end{cases} \quad (14)$$

Therefore, solutions of Expression (1) can be expressed by $\varphi_n(\vec{x})$. For $\forall u(\vec{x}) \in V(\Omega)$, $u(\vec{x})$ can be presented as a series of eigenfunctions satisfying absolute and uniform convergence, i.e.

$$u(x) = \sum_{n=1}^{+\infty} c_n \varphi_n(x), \quad c_n = \langle u, \varphi_n \rangle \quad (15)$$

Expression (15) has profound physical meaning. $\sqrt{\lambda_n}$ and $\varphi_n(\vec{x})$ are the $n-th$ order structural natural frequency and the $n-th$ order vibration mode. If $u(\vec{x})$ is regarded as a vibration amplitude function, it can be decomposed into a superposition of vibration modes at each order natural frequency, where the coefficient $c_n$ is the vibration magnitude at $\varphi_n(\vec{x})$. This is equivalent to spectral decomposition. Imagine such a scene. When we enjoy a piece a music, our brains constantly decompose the instantaneous vibration amplitude $u(\vec{x})$ according to Expression (15), and meanwhile, perceive the vibration coefficients $c_n$ and mark them with $\sqrt{\lambda_n}$. For a musical instrument, $\sqrt{\lambda_1}$ is its fundamental frequency (tone) and the remaining eigenvalues are overtones. Different musical instruments have different vibration patterns, and their eigenfunctions $\{\varphi_n(\vec{x}), n = 1, 2 \cdots\}$ are also different. However, after tuning the tone of different musical instruments, their fundamental frequency of each note is consistent.

The eigenfunctions and eigenvalues satisfy Expression (13), so we have

$$L[\varphi_n(\vec{x})] = \lambda_n \varphi_n(\vec{x}) \quad (16)$$

Multiplying both sides of Expression (16) by $\varphi_n(\vec{x})$ and integrating by parts, we can yield

$$\langle \varphi_n, \varphi_m \rangle_U = \begin{cases} \lambda_n & m = n \\ 0 & m \neq n \end{cases} \quad (17)$$

Expression (17) shows that the eigenfunctions also satisfy the orthogonal condition with respect to the mutual-energy inner product. So, these eigenfunctions $\{\varphi_n(\vec{x})/\sqrt{\lambda_n}, n = 1, 2 \cdots\}$ can also be used as basis functions to span the mutual-energy inner product space $V(\Omega)$.

Substituting Expression (15) into Expression (5) and applying Expression (17), we have

$$\langle u, u \rangle_U = \sum_{n=1}^{\infty} c_n^2 \lambda_n \quad (18)$$

If $u(\vec{x})$ satisfies the normalization condition $\|u\|_2^2 = 1$ or $\sum_{n=1}^{\infty} c_n^2 = 1$, based on the Expressions (14, 18), the eigenvalue $\lambda_n$ satisfies

$$\begin{aligned}
\lambda_n = \min_{u \in V(\Omega)} &\langle u, u \rangle_U \\
\text{s.t.} \quad & \\
& \langle u, u \rangle = 1 \\
& \langle u, \varphi_i \rangle = 0 \quad i = 1, 2 \cdots n-1
\end{aligned} \quad (19)$$

where the optimal solution of $u(\vec{x})$ is the eigenfunction $\varphi_n(\vec{x})$.

Similarly, the deformation $v(\vec{x})$ caused by $g(\vec{x})$ can be expressed as

$$v(\vec{x}) = \sum_{n=1}^{\infty} d_n \varphi_n(\vec{x}) \quad (20)$$

Where, $d_n$ is the amplitude coefficient and can be interpreted as the component of $v(\vec{x})$ at the $n-th$ vibration mode $\varphi_n(\vec{x})$. Substituting Expression (20) into Expression (12) and using the orthogonal condition (17), we have

$$\min_{d_n, n=1,2\cdots\infty} \pi[d_1, d_2 \cdots] = \sum_{n=1}^{\infty} \left( \frac{1}{2} \lambda_n d_n^2 - g_n d_n \right) \quad (21)$$

Where, the coefficient $g_n$ is the projection of $g(\vec{x})$ on $\varphi_n(\vec{x})$ with respect to the Euclidean inner product

$$g_n = \langle g, \varphi_n \rangle \quad (22)$$

Enforcing the derivative of $\pi[d_1, d_2 \cdots]$ in Expression (21) with respect to $d_n$ to zero, we have

$$d_n = \frac{g_n}{\lambda_n} \quad n = 1, 2 \cdots \infty \quad (23)$$

According to the series representation of $u(\vec{x})$ in Expression (15), if $u(\vec{x})$ is the deformation caused by $f(\vec{x})$, the coefficient $c_n$ satisfies

$$c_n = \frac{f_n}{\lambda_n}, \quad f_n = \langle f, \varphi_n \rangle \quad n = 1, 2 \cdots \infty \quad (24)$$

Substituting the Expressions (15, 20, 23, 24) into Expression (5) and using the orthogonal condition (17), we have

$$\langle u, v \rangle_U = \sum_{n=1}^{\infty} \frac{f_n \cdot g_n}{\lambda_n} \quad (25)$$

Generally speaking, the external force $f(\vec{x}) \notin V(\Omega)$, because $f(\vec{x})$ does not satisfy the homogeneous boundary conditions, i.e. $l[f(\vec{x})] \neq 0$. In this case, $\sum_{n=1}^{\infty} f_n \varphi_n(\vec{x})$ is equal to the projection of $f(\vec{x})$ on $V(\Omega)$ or the optimal approximation of $f(\vec{x})$ in $V(\Omega)$. Of course, in order to make $f(\vec{x}) \in V(\Omega)$, we may expand the design domain and simplify the boundary condition. For example, after expanding the design domain, set a fixed boundary and let $\sigma(\vec{x}) = \infty$, or set a mirror boundary and let $\sigma(\vec{x}) = 0$. In these cases, $l[f(\vec{x})] = 0$ and $f(\vec{x}) = \sum_{n=1}^{\infty} f_n \varphi_n(\vec{x})$. Then applying the orthogonal condition (14) yields

$$\langle f, g \rangle = \sum_{n=1}^{\infty} f_n \cdot g_n \quad (26)$$

After $f(\vec{x})$ and $g(\vec{x})$ are expressed as a superposition of eigenfunctions of the operator $L$, through comparing the mutual-energy inner product $\langle u,v \rangle_U$ in Expression (25) and the Euclidean inner product $\langle f,g \rangle$ in Expression (26), it can be found that, the mutual-energy inner product has an advantage of enhancing low-frequency coordinate components ($\sqrt{\lambda_n} < 1$) and suppressing high-frequency coordinate components ($\sqrt{\lambda_n} > 1$). In other words, if $f(\vec{x})$ and $g(\vec{x})$ are regarded as signals, the mutual-energy inner product can augment low-frequency eigenfunction components and filter out high-frequency eigenfunction components of the signals, with the help of a structural model.

## 4. Mutual-Energy Inner Product Optimization Model for Feature Extraction

Assume that $D = \{(X^{(i)}(\vec{x}), y^{(i)})\}_{i=1}^{N}$ is a training dataset with $N$ samples, and each sample is represented as $X^{(i)}(\vec{x})$, while $y^{(i)}$ represents class labels. For example, samples are divided into 2 classes, and $D$ includes 2 subsets $D_1 = \{X^{(i)}(x) | (X^{(i)}(\vec{x}) \in D, y^{(i)} = 1)\}_{i=1}^{N_1}$ and $D_0 = \{X^{(i)}(x) | (X^{(i)}(\vec{x}) \in D, y^{(i)} = 0)\}_{i=1}^{N_0}$, where $N = N_1 + N_0$. Generally, samples in different classes are assumed to be random variables, which are independent and have identical distributions.

We hope to find an appropriate feature coordinate system to represent $X^{(i)}(\vec{x}) \in D$ and use fewer coordinate components to classify samples. If there is no further information, we may select the means of probability distribution of $D_1$ and $D_0$ as reference features. In order to design a feature extraction model, two points should be considered: one to enhance feature information, and the other to suppress the effect of random noise. We resort to a structural model and use the mutual-energy inner product to extract features. Its main idea is to map the data features to a low-frequency eigenfunction space of the structural model.

If $f(\vec{x})$ and $g(\vec{x})$ are used to represent the means of the probability distribution of $D_1$ and $D_0$, respectively, their unbiased estimates can be written as

$$f(\vec{x}) = \frac{1}{N_1} \sum_{X^{(i)}(\vec{x}) \in D_1} X^{(i)}(\vec{x}), \quad g(\vec{x}) = \frac{1}{N_0} \sum_{X^{(i)}(\vec{x}) \in D_0} X^{(i)}(\vec{x}) \tag{27}$$

We regard $X^{(i)}(\vec{x})$, $f(\vec{x})$ and $g(\vec{x})$ as external forces acting on the structural model, and use $d^{(i)}(\vec{x})$, $u(\vec{x})$ and $v(\vec{x})$ to represent their corresponding deformations, respectively. If we represent the selected reference feature in $V(\Omega)$ as $\alpha(\vec{x})$, we can use the mutual-energy inner product $\langle d^{(i)}, \alpha \rangle_U$ to extract the feature coordinate component of $X^{(i)}(\vec{x})$. In order to construct the feature extraction optimization model, we first select $u(\vec{x})$ as the reference feature $\alpha(\vec{x})$ and try to explore physical meanings of the structural model when $\langle d^{(i)}, u \rangle_U$ is maximum, minimum or equal to zero.

In order to enhance the feature information of samples in $D_1$, a high statistical mean value $\mu_1$ should be given

$$\mu_1 = \frac{1}{N_1} \sum_{X^{(i)}(\vec{x}) \in D_1} \langle d^{(i)}, u \rangle_U = \left\langle \frac{1}{N_1} \sum_{X^{(i)}(\vec{x}) \in D_1} d^{(i)}, u \right\rangle_U = \langle u, u \rangle_U \tag{28}$$

With a primary objective

$$\max_{p(\vec{x}), q(\vec{x})} \mu_1 = \langle u, u \rangle_U \tag{29}$$

In Expression (29), the mutual-energy inner product and deformations are functions of $p(\vec{x})$ and $q(\vec{x})$, and its physical meaning is not intuitive. So next, we will conduct quantitative analysis to reveal structural characteristics hidden in Expression (29).

According to the minimum energy principle (12), if an optimal solution of $u(\vec{x})$ is obtained, the derivative of the objective at the optimal solution in any direction $\delta u(\vec{x})$ is zero, satisfying

$$\frac{d}{d\tau}\pi[u(\vec{x})+\tau\cdot\delta u(\vec{x})]=0 \qquad \forall \delta u(\vec{x})\in C(\Omega) \tag{30}$$

Through calculating Expression (30), we obtain the relationship between $u(\vec{x})$ and $f(\vec{x})$

$$\langle u,\delta u\rangle_U - \langle f,\delta u\rangle = 0 \qquad \forall \delta u(\vec{x})\in C(\Omega) \tag{31}$$

Expression (31) is a structural static equilibrium equation, and is also a constraint on $u(\vec{x})$ in optimization problem (29). In Expression (31), letting $\delta u(\vec{x}) = u(\vec{x})$ yields

$$\langle u,u\rangle_U = \langle f,u\rangle \tag{32}$$

Substituting Expression (32) into Expression (12) yields the optimal value $\pi[u]$ of the objective

$$\pi[u] = -\frac{1}{2}\langle u,u\rangle_U \tag{33}$$

Through substituting the Expressions (12, 33) into the optimization problem (29), Expression (29) is transformed into an unconstrained optimization problem

$$\max_{p(\vec{x}),q(\vec{x}),u(\vec{x})} \mu_1[u,p,q] = 2\langle f,u\rangle - \langle u,u\rangle_U \tag{34}$$

If $p(\vec{x})$ and $q(\vec{x})$ are given, $\mu_1[u,p,q]$ in Expression (34) is a quadratic and concave functional with respect to $u(\vec{x})$, due to $p(\vec{x}) > 0$ and $q(\vec{x}) > 0$. If $u(\vec{x})$ is given, $\mu_1[u,p,q]$ is a linear functional with respect to $p(\vec{x})$ and $q(\vec{x})$. Through using the Univariate Search Method to solve Expression (34), if $p(\vec{x})$ and $q(\vec{x})$ are given, the maximum value of $\mu_1[u,p,q]$ can be found by solving Expression (31) for $u(\vec{x})$, and if $u(\vec{x})$ is given, the maximum value of $\mu_1[u,p,q]$ will be reached on the lower bounds of $p(\vec{x})$ and $q(\vec{x})$. So, the lower bounds of $p(\vec{x})$ and $q(\vec{x})$ must be larger than zero to ensure that Expression (29) has a finite optimal solution. In addition, the upper bounds of $p(\vec{x})$ and $q(\vec{x})$ should also be constrained to avoid the trivial solution $u(\vec{x}) = 0$. Therefore, when the optimization objective is to maximize the mutual-energy inner product, as shown in Expression (29), its optimal structural model would be the minimum stiffness structure, and the selected feature belongs to a low-frequency eigenfunction subspace. On the contrary, if the optimization objective is to minimize the mutual-energy inner product, the optimal structural model would be the maximum stiffness, and the selected feature would be mapped to a high-frequency eigenfunction subspace.

In addition, when using the mutual-energy inner product to extract feature information $f(\vec{x})$ of samples in $D_1$, the feature information $g(\vec{x})$ of samples in $D_0$ should be suppressed. So, a small statistical mean value $\mu_0$ is given

$$\mu_0 = \frac{1}{N_0}\sum_{X^{(i)}(\vec{x})\in D_0}\langle d^{(i)},u\rangle_U = \left\langle \frac{1}{N_0}\sum_{X^{(i)}(\vec{x})\in D_0} d^{(i)},u\right\rangle_U = \langle v,u\rangle_U \tag{35}$$

Here we may set $\mu_0$ to be zero or even negative, and impose constraints on the structural model

$$\langle v,u\rangle_U \leq 0 \tag{36}$$

In Expression (31), setting $\delta u(\vec{x}) = v(\vec{x})$ yields $\langle u,v\rangle_U = \langle f,v\rangle$. Replacing $f(\vec{x})$ with $g(\vec{x})$, and exchanging $u(\vec{x})$, $v(\vec{x})$, we have

$$\langle u,v\rangle_U = \langle f,v\rangle = \langle g,u\rangle \tag{37}$$

If Expression (36) satisfies $\langle v,u\rangle_U = 0$, then $u(\vec{x})$ and $v(\vec{x})$ are required to be orthogonal with respect to the mutual-energy inner product. Although the means of two classes of samples are generally not orthogonal in the continuous function space $C(\Omega)$, i.e. $\langle f,g\rangle \neq 0$, the orthogonality of $u(\vec{x})$ and $v(\vec{x})$ can be easily realized according to Expression (37). For example, if setting $p(\vec{x}) = 0$ and dividing the domain $\Omega$ into two sub-regions according to same or opposite signs of $f(\vec{x})$ and $g(\vec{x})$, we can adjust $q(\vec{x})$ in the two sub-regions and control positive and negative work done by external forces $g(\vec{x})$ on deformations $u(\vec{x})$, so as to make the total work $\langle g,u \rangle$ in Expression (37) zero. According to Expression (25), this can also be understood as designing a structural model and adjusting its eigenfunctions and eigenvalues, so as to use these eigenvalues as weights to achieve weighted orthogonality of $f(\vec{x})$ and $g(\vec{x})$. Further, $\langle v,u \rangle_U \leq 0$ can be regarded as a relaxation of orthogonal constraints on the mutual-energy inner product, which can be realized by adjusting $p(\vec{x})$ and $q(\vec{x})$ to make $\langle g,u \rangle < 0$. Geometrically, this means that the angle between $u(\vec{x})$ and $v(\vec{x})$ in the mutual-energy inner product space $V(\Omega)$ is not an acute angle. If $\mu_0$ is required to be minimal

$$\min_{p(\vec{x}),q(\vec{x})} \quad \mu_0 = \langle v,u \rangle_U \tag{38}$$

based on Expression (12), similar to the discussion on Expression (29), the optimization problem (38) can be transformed into an unconstrained form

$$\min_{u(\vec{x}),v(\vec{x}),p(\vec{x}),q(\vec{x})} \max_{z(\vec{x})} \quad \mu_0[z,u,v,p,q] \tag{39}$$

Where, $z(\vec{x}) \in C(\Omega)$ is a slack variable introduced to relax the constraint, which is the constraint of the static equilibrium equation describing structural deformation due to $f(\vec{x})$ and $g(\vec{x})$ acting on the structure simultaneously. The objective can be expressed as

$$\mu_0[z,u,v,p,q] = \frac{1}{2}\langle u,u \rangle_U + \frac{1}{2}\langle v,v \rangle_U - \frac{1}{2}\langle z,z \rangle_U - \langle f,u-z \rangle - \langle g,v-z \rangle \tag{40}$$

Obviously, if $p(\vec{x}) > 0$ and $q(\vec{x}) > 0$ are given, $\mu_0[z,u,v,p,q]$ is a quadratic functional of $u(\vec{x})$, $v(\vec{x})$, $z(\vec{x})$. $\mu_0[z,u,v,p,q]$ is convex with respect to $u(\vec{x})$, $v(\vec{x})$, and is concave with respect to $z(\vec{x})$. If $u(\vec{x})$, $v(\vec{x})$ and $z(\vec{x})$ are given, $\mu_0[z,u,v,p,q]$ is linear with respect to $p(\vec{x})$ and $q(\vec{x})$.

In order to design a feature coordinate to classify samples in $D$, the objective is to maximize $\mu_1 - \mu_0$ first. By combining the Expressions (28, 35), the optimization objective can be expressed as

$$\min_{p(\vec{x}),q(\vec{x})} \quad \mu_0 - \mu_1 = \langle v-u,u \rangle_U \tag{41}$$

Then, to improve classification accuracy, distributions of samples in $D_1$ and $D_0$ along the feature coordinate $u(\vec{x})$ should also be considered, and their variances should be small. The variances of $D_1$ and $D_0$ are high-order functions of $u(\vec{x})$, $v(\vec{x})$, $p(\vec{x})$ and $q(\vec{x})$, so putting them into the optimization objective function (41) will destroy its low-order characteristics.

In order to improve computational efficiency, the sum of the absolute values of sample deviations from the mean are used to replace variances, and only some samples in $D_1$ and $D_0$ are selected to calculate. In the subset $D_1$, we only select $M_1$ samples $S_1 = \{X^{(i)}(\vec{x}) | X^{(i)}(\vec{x}) \in D_1, \langle X^{(i)},u \rangle < \mu_1\}_{i=0}^{M_1}$, whose components on $u(\vec{x})$ are less than $\mu_1$, and calculate their mean absolute deviation $\delta_1$. In the subset $D_0$, we only select $M_0$ samples $S_0 = \{X^{(i)}(\vec{x}) | X^{(i)}(\vec{x}) \in D_0, \langle X^{(i)},u \rangle > \mu_0\}_{i=0}^{M_0}$, whose components on $u(\vec{x})$ are larger than $\mu_0$, and calculate their mean absolute deviation $\delta_0$. $\delta_1$ and $\delta_0$ can be expressed as

$$\begin{aligned} \delta_1 &= \frac{1}{M_1} \sum_{X^{(i)}(\vec{x}) \in S_1} \langle f - X^{(i)}, u \rangle = \langle u,u \rangle_U - \left\langle \frac{1}{M_1} \sum_{X^{(i)}(\vec{x}) \in S_1} X^{(i)}, u \right\rangle \\ \delta_0 &= \frac{1}{M_0} \sum_{X^{(i)}(\vec{x}) \in S_0} \langle X^{(i)} - g, u \rangle = \left\langle \frac{1}{M_0} \sum_{X^{(i)}(\vec{x}) \in S_0} X^{(i)}, u \right\rangle - \langle v,u \rangle_U \end{aligned} \tag{42}$$

Through using the Expressions (42, 41), and considering the means and the mean absolute deviations of the samples, the optimization objective can be written as

$$\min_{p(\vec{x}),q(\vec{x})} \quad J[p(\vec{x}),q(\vec{x})] = \lambda(\mu_0 - \mu_1) + (1-\lambda)(\delta_0 + \delta_1) \tag{43}$$

Where $\lambda$ is a weight variable, satisfying $0 \leq \lambda \leq 1$. To simplify Expression (42), an auxiliary deformation function $w(\vec{x}) \in V(\Omega)$ is defined as

$$\langle w, \delta w \rangle_U - \langle h, \delta w \rangle = 0 \quad \forall \delta w(\vec{x}) \in C(\Omega) \tag{44}$$

Where, $h(\vec{x})$ can be regarded as an external force corresponding to $w(\vec{x})$, satisfying

$$h(\vec{x}) = \frac{1}{M_0} \sum_{X^{(i)}(\vec{x}) \in S_0} X^{(i)}(\vec{x}) - \frac{1}{M_1} \sum_{X^{(i)}(\vec{x}) \in S_1} X^{(i)}(\vec{x}) \tag{45}$$

By substituting the Expressions (41, 42, 44, 45) into Expression (43), the optimization objective is simplified as

$$\min_{p(\vec{x}),q(\vec{x})} \quad J[p(\vec{x}),q(\vec{x})] = \langle c, u \rangle_U \tag{46}$$

Here, $c(\vec{x}) \in V(\Omega)$ is a combination of deformation functions, and can be expressed as

$$c(\vec{x}) = (1-2\lambda)(u(\vec{x}) - v(\vec{x})) + (1-\lambda)w(\vec{x}) \tag{47}$$

In order to improve the generalization of the data classifier, regularizers should be added to the optimization model. Here, $\|p\|_1$ and $\|q\|_1$ stand for the 1-norms of $p(\vec{x})$ and $q(\vec{x})$, respectively, and are used as regularizers to avoid increasing the order of the optimization model. Meanwhile, these regularizers are treated as two constraints by directly setting the values of $\|p\|_1$ and $\|q\|_1$. Due to $p(\vec{x}) > 0$ and $q(\vec{x}) > 0$, $\|p\|_1$ and $\|q\|_1$ can be simply written as

$$\|p(\vec{x})\|_1 = \int_\Omega p(\vec{x}) d\Omega, \quad \|q(\vec{x})\|_1 = \int_\Omega q(\vec{x}) d\Omega \tag{48}$$

It should be noted that objective (46) is built by taking the mean $f(\vec{x})$ of $D_1$ as the reference feature and selecting the deformation $u(\vec{x})$ as the reference feature coordinate axis. If other deformation functions $\alpha(\vec{x})$ are selected as the reference feature coordinate axis, the results are similar. For example, $\alpha(\vec{x})$ can be set as $u(\vec{x})$, $v(\vec{x})$, $u(\vec{x}) - v(\vec{x})$, or others. Through setting $\alpha(\vec{x})$ as the reference feature coordinate axis, the optimization model can be summarized as

$$\begin{aligned}
&\min_{p(\vec{x}),q(\vec{x})} \quad J[p,q] = \langle c, \alpha \rangle_U \\
&s.t. \\
&\langle u, \delta u \rangle_U = \langle f, \delta u \rangle, \quad \langle v, \delta v \rangle_U = \langle g, \delta v \rangle, \quad \langle w, \delta w \rangle_U = \langle h, \delta w \rangle \\
&\langle u, v \rangle_U \leq 0 \\
&\|p\|_1 = Tolp, \quad \|q\|_1 = Tolq \\
&p_{\min} \leq p(\vec{x}) \quad q_{\min} \leq q(\vec{x})
\end{aligned} \tag{49}$$

Where $\delta u$, $\delta v$ and $\delta w$ are arbitrary continuous functions on $\Omega$; $p_{\min} > 0$ and $q_{\min} > 0$, are lower bounds of $p(\vec{x})$ and $q(\vec{x})$; $Tolp$ and $Tolq$ are two constants; $c(\vec{x})$, $f(\vec{x})$, $g(\vec{x})$ and $h(\vec{x})$ are given in the Expressions (47, 27, 45). $S_1$ and $S_0$ should be determined according to the reference feature coordinate axis, and can be rewritten as

$$\begin{aligned}
S_1 &= \left\{ X^{(i)}(\vec{x}) \mid X^{(i)}(\vec{x}) \in D_1, \langle X^{(i)}, \alpha \rangle < \langle f, \alpha \rangle \right\}_{i=0}^{M_1} \\
S_0 &= \left\{ X^{(i)}(\vec{x}) \mid X^{(i)}(\vec{x}) \in D_0, \langle X^{(i)}, \alpha \rangle > \langle g, \alpha \rangle \right\}_{i=0}^{M_0}
\end{aligned} \tag{50}$$

## 5. Mutual-Energy Inner Product Feature Coordinate Optimization Algorithm

The EFM is used to solve the differential equation (1) to realize the mapping from $f(\vec{x}), g(\vec{x}), h(\vec{x})$ to $u(\vec{x}), v(\vec{x}), w(\vec{x})$ in the optimization model (49). We divide the domain $\Omega$ into $Ne$ elements $\Omega_s^{(e)}$ $(e=1,2\cdots Ne)$, and assume the $e-th$ element $\Omega_s^{(e)}$ has $Nd$ nodes. For the $i-th$ $(i=1,2\cdots Nd)$ node in $\Omega_s^{(e)}$, its global coordinate in $\Omega$, deformation value $u(\vec{x}_i^{(e)})$ and interpolation basis function are denoted as $\vec{x}_i^{(e)}$, $u_i^{(e)}$, $N_i(\vec{\xi})$, respectively, where $\vec{\xi} \in R^n$, is the local coordinate of the element $\Omega_s^{(e)}$. In this way, for an element, its global and local coordinate relationship $\vec{x}^{(e)}(\vec{\xi})$ and the element deformation function $u^{(e)}(\vec{\xi})$ can be expressed as [37]

$$\vec{x}^{(e)}(\vec{\xi}) = \sum_{j=1}^{Nd} N_j(\vec{\xi})\vec{x}_j^{(e)}, \quad u^{(e)}(\vec{\xi}) = \sum_{j=1}^{Nd} N_j(\vec{\xi})u_j^{(e)} \tag{51}$$

It is assumed that $\vec{N}$ is an $Nd$-dimensional row vector with the $j-th$ component $N_j(\vec{\xi})$; $L$ is an $n \times Nd$ matrix with the entry $L_{ij} = \partial N_j(\vec{\xi})/\partial \xi_i$, where $\xi_i$ is the $i-th$ component of the local coordinate $\vec{\xi}$; and $X$ is an $Nd \times n$ matrix with the entry $X_{ij} = x_{ij}^{(e)}$, where $x_{ij}^{(e)}$ is the $j-th$ component of the element node coordinates $\vec{x}_i^{(e)}$. Applying Expression (51), the $n \times n$ Jacobi matrix $J$ for the transformation between global and local coordinates, the deformation function $u^{(e)}(\vec{x})$ and its $n$-dimensional gradient vector $\nabla u^{(e)}(\vec{x}) = [\partial u^{(e)}/\partial x_1 \quad \partial u^{(e)}/\partial x_2 \quad \cdots \quad \partial u^{(e)}/\partial x_n]^T$, can be expressed in the concise and compact form

$$u^{(e)}(\vec{x}) = \vec{N} \cdot \vec{u}^{(e)}, \quad \nabla u^{(e)}(\vec{x}) = B \cdot \vec{u}^{(e)}, \quad J = L \cdot X, \quad B = J^{-1}L \tag{52}$$

Where, $\vec{u}^{(e)} = [u_1^{(e)} \quad u_2^{(e)} \quad \cdots \quad u_{Nd}^{(e)}]^T$ is a vector with the component $u_i^{(e)}$, which is the deformation value of the $i-th$ node in the $e-th$ element, and $B$ is an $n \times Nd$ matrix. In the optimization model (49), the design variables are $p(\vec{x})$ and $q(\vec{x})$. We assume that $p(\vec{x})$ and $q(\vec{x})$ in each element are constants $p_e$ and $q_e$. So the design variables can be expressed as $\vec{p} = [p_1 \quad p_2 \quad \cdots \quad p_{Ne}]^T$ and $\vec{q} = [q_1 \quad q_2 \quad \cdots \quad q_{Ne}]^T$ in $\Omega$.

Substituting Expression (52) into the mutual-energy expressions (5, 9) yields

$$\langle u,u \rangle_U = \sum_{e=1}^{Ne} \vec{u}^{(e)T} K_s^{(e)} \vec{u}^{(e)}, \quad \langle f,u \rangle = \sum_{e=1}^{Ne} \vec{u}^{(e)T} \vec{f}^{(e)} \tag{53}$$

Where $K_s^{(e)}$ is an $Nd \times Nd$ element stiffness matrix, which is a positive semidefinite symmetric matrix and can be expressed as

$$K_s^{(e)} = p_e K_p^{(e)} + q_e K_q^{(e)} + K_\sigma^{(e)} \tag{54}$$

In Expression (54), $K_s^{(e)}$ is a linear function of $p_e$ and $q_e$; $K_p^{(e)}$ and $K_q^{(e)}$ are corresponding coefficient matrices; and $K_\sigma^{(e)}$ is the contribution of the boundary constraint to the element stiffness matrix. If the element boundary does not overlap with the design domain boundary, then $K_\sigma^{(e)} = 0$. Here, $K_p^{(e)}$, $K_q^{(e)}$, $K_\sigma^{(e)}$ can be calculated by

$$K_p^{(e)} = \int_{\Omega_s^{(e)}} B^T B d\Omega, \quad K_q^{(e)} = \int_{\Omega_s^{(e)}} \vec{N}^T \vec{N} d\Omega, \quad K_\sigma^{(e)} = \int_{\Gamma_s^{(e)}} \sigma(\vec{x}) \vec{N}^T \vec{N} d\Omega \tag{55}$$

In Expression (53), $\vec{f}^{(e)}$ is the equivalent node input vector, resulting from the equivalent action between the force $f(\vec{x})$ on the element and the force $\vec{f}^{(e)}$ on the node, and satisfies

$$\vec{f}^{(e)} = \int_{\Omega_s^{(e)}} f(\vec{x}) \vec{N}^T d\Omega \tag{56}$$

It is assumed that the design domain $\Omega$ comprises $M$ element nodes. We number these nodes globally, and use two $M$-dimension vectors $\vec{u}$ and $\vec{f}$ to denote values of $u(\vec{x})$ and $f(\vec{x})$ at all the nodes. The components of $\vec{u}$ and $\vec{f}$ are $u_i$ and $f_i$, where the subscript $i$ is the global node number. The component $f_i$ can be calculated through Expression (56). Expression (56) is calculated for each element adjacent to the $i-th$ global node, and $f_i$ is the superposition of the element node corresponding to the $i-th$ global node.

Based on the relationship between the local and global node numbers, Expression (53) can be rewritten as

$$\langle u,u\rangle_U = \vec{u}^T K \vec{u} = \sum_{e=1}^{Ne} \vec{u}^{(e)T} K_s^{(e)} \vec{u}^{(e)}, \qquad \langle f,u\rangle = \vec{u}^T \vec{f} = \sum_{e=1}^{Ne} \vec{u}^{(e)T} \vec{f}^{(e)} \qquad (57)$$

Where $K$ is the global stiffness matrix, an $M \times M$ positive definite symmetric matrix. Substituting Expression (57) into Expression (12) yields

$$\min_{\vec{u}} \quad \pi[\vec{u}] = \frac{1}{2}\vec{u}^T K \vec{u} - \vec{u}^T \vec{f} \qquad (58)$$

Based on Expression (58), the solution of the differential equation (1) satisfies

$$K\vec{u} - \vec{f} = 0 \qquad (59)$$

Similarly, assume that the input of Expression (1) is $g(\vec{x})$ and the corresponding solution is $v(\vec{x})$; $\vec{v}$ is the global node vector corresponding to $v(\vec{x})$ on $\Omega$, and $\vec{v}^{(e)}$ is the element node vector corresponding to $v(\vec{x})$ on $\Omega_s^{(e)}$; and $\vec{g}$ is the equivalent node input vector corresponding to $g(\vec{x})$. We have

$$K\vec{v} - \vec{g} = 0 \qquad (60)$$

Similar to the derivation of Expression (57), through using the Expressions (59, 60), the mutual-energy expression of $u(\vec{x})$ and $v(\vec{x})$ can be derived

$$\langle u,v\rangle_U = \vec{u}^T K \vec{v} = \sum_{e=1}^{Ne} \vec{u}^{(e)T} K_s^{(e)} \vec{v}^{(e)}, \qquad \langle u,v\rangle_U = \vec{u}^T \vec{f} = \vec{v}^T \vec{g} \qquad (61)$$

In Expression (61), the first equation is used for model optimization, and the second equation is used for data classifier training and prediction, avoiding the need to solve for the Expressions (59, 60).

After discretizing the design domain by finite elements, the differential equation (1) is converted into a system of linear equations, and the mutual-energy definition (5) can be expressed by matrix and vector product. In this way, the optimization model (49) can be rewritten in the vector form

$$\min_{\vec{p},\vec{q}} \quad J[\vec{p},\vec{q}] = \vec{c}^T K \vec{\alpha}$$

s.t.
$$\begin{aligned} K\vec{u} = \vec{f}, \qquad K\vec{v} = \vec{g}, \qquad K\vec{w} = \vec{h} \\ G[\vec{p},\vec{q}] = \vec{u}^T K \vec{v} \leq 0 \\ \sum_{e=1}^{Ne} p_e = Tolp, \quad \sum_{e=1}^{Ne} q_e = Tolq \\ p_{\min} \leq p_e, \quad q_{\min} \leq q_e, \quad e = 1,2\cdots Ne \end{aligned} \qquad (62)$$

Here, $\vec{\alpha}$ is the finite element node vector corresponding to the selected reference feature coordinate, and can be statistical features of sample sets, or their combination, for example,

$$\vec{\alpha} = \vec{u} \quad or \quad \vec{\alpha} = \vec{v} \quad or \quad \vec{\alpha} = \vec{u} - \vec{v} \qquad (63)$$

Meanwhile, $\vec{f}$, $\vec{g}$, $\vec{h}$ are the finite element node vectors corresponding to the mean and deviation of samples, and $\vec{c}$ is the temporary node vector generated by the mean and deviation. Expression (47) can be rewritten as

$$\vec{c} = (1-2\lambda)(\vec{u}-\vec{v}) + (1-\lambda)\vec{w} \qquad (64)$$

The significant advantage of the optimization model (62) is that, $K$ is a positive definite symmetric matrix and is linear with respect to design variables $\vec{p}$ and $\vec{q}$, and meanwhile, coefficient matrices corresponding to components of design variables are positive semidefinite matrices, convenient for algorithm design. Intermediate variables $\vec{u}$, $\vec{v}$, $\vec{w}$ are functions of design variables and can be calculated by using the linear equations, and the optimization model (62) can be solved by the sequential linearization algorithm. The objective $J[\vec{p},\vec{q}]$ and the constraint $G[\vec{p},\vec{q}]$ are nonlinear, and their derivatives with respect to design variables need to be calculated. The derivative of $G[\vec{p},\vec{q}]$ with respect to $p_e$ is

$$\frac{\partial G}{\partial p_e} = \frac{\partial \vec{u}^T}{\partial p_e} K \vec{v} + \vec{u}^T \frac{\partial K}{\partial p_e} \vec{v} + \vec{u}^T K \frac{\partial \vec{v}}{\partial p_e} \qquad e = 1,2\cdots Ne \tag{65}$$

Where $\dfrac{\partial \vec{u}}{\partial p_e}$ and $\dfrac{\partial \vec{v}}{\partial p_e}$ are determined by taking the derivative of $K\vec{u} = \vec{f}$ and $K\vec{v} = \vec{g}$ with respect to $p_e$

$$\frac{\partial K}{\partial p_e}\vec{u} + K\frac{\partial \vec{u}}{\partial p_e} = 0, \qquad \frac{\partial K}{\partial p_e}\vec{v} + K\frac{\partial \vec{v}}{\partial p_e} = 0 \tag{66}$$

Substituting Expression (66) into Expression (44) yields

$$\frac{\partial G}{\partial p_e} = -\vec{u}^T \frac{\partial K}{\partial p_e} \vec{v} \tag{67}$$

Substituting Expression (54) into Expression (67) yields $\partial G/\partial p_e$. Similarly, $\partial G/\partial q_e$ can also be computed

$$\frac{\partial G}{\partial p_e} = -\vec{u}^{(e)T} K_p^{(e)} \vec{v}^{(e)}, \qquad \frac{\partial G}{\partial q_e} = -\vec{u}^{(e)T} K_q^{(e)} \vec{v}^{(e)} \tag{68}$$

The Expressions (63, 64) show that $\vec{c}$ and $\vec{\alpha}$ are linear combinations of $\vec{u}$, $\vec{v}$, $\vec{w}$. According to the superposition principle, $\partial \vec{c}/\partial p_e$ and $\partial \vec{\alpha}/\partial p_e$ also satisfy equations similar to Expression (66), and have exactly the same derivation as Expression (68). So, we get

$$\frac{\partial J}{\partial p_e} = -\vec{c}^{(e)T} K_p^{(e)} \vec{\alpha}^{(e)}, \qquad \frac{\partial J}{\partial q_e} = -\vec{c}^{(e)T} K_q^{(e)} \vec{\alpha}^{(e)} \tag{69}$$

**Optimization Algorithm I: Mutual-energy inner product feature coordinate optimization algorithm**

Based on the Expressions (68, 69), the optimization model (62) can be solved by the sequential linearization algorithm. The algorithm steps are summarized as follows.

(1) Use vectors to represent sample data

Convert sample data $X^{(i)}(\vec{x})$ in the training subsets $D_1$ and $D_0$ into the finite element node vectors $\vec{X}^{(i)} \in R^M$. Based on Expression (70), first calculate the element node vectors $\vec{X}^{(i)(e)} \in R^{Nd}$, and then use them to assemble the global node vector $\vec{X}^{(i)}$.

$$\vec{X}^{(i)(e)} = \int_{\Omega_s^{(e)}} X^{(i)}(\vec{x}) \vec{N}^T d\Omega \qquad e = 1,2\cdots Ne \tag{70}$$

(2) Set optimization constants and initial values of the design variables

① Set optimization constants

Set $\lambda$, the weight of the mean and deviation, with the requirement $\lambda \in [0,1]$; set the total amount $Tolp$, $Tolq$ and lower bounds $p_{\min}$, $q_{\min}$ of the design variables; set the moving limit $\Delta x_{\max}$ of the design variables for the linear programming; set the design variable minimum increment $\varepsilon_x$ and the objective function minimum increment $\varepsilon_J$, which are used to determine if the optimization ends or not.

② Set initial values of the design variables

Set $p_e = p_e^{(0)}$, $q_e = q_e^{(0)}$ $(e = 1,2\cdots Ne)$. Generally, set $p_e^{(0)} = Tolp/M$, $q_e^{(0)} = Tolq/M$.

(3) Calculate the current value of the objective function

① Calculate the element stiffness matrices and assemble the global stiffness matrix

Based on the Expressions (54, 55), calculate the element stiffness matrices $K_s^{(e)}$ $(e = 1,2\cdots Ne)$.

The element stiffness matrix is linear with respect to $p_e$ and $q_e$, and coefficient matrices are determined only by element interpolation basis functions, so the calculation can be done prior to the optimization to speed up the optimization process. Then, assemble the global stiffness matrix according to node numbers. Since $K$ is a positive definite symmetric matrix, through performing

Cholesky decomposition on it, we can have $K = L \cdot L^T$, where $L$ is a lower triangular matrix.

② Compute the mean vectors $\vec{u}$ and $\vec{v}$, and select the reference feature coordinate axis $\vec{\alpha}$

$$\begin{cases} \vec{u} = L^{-T}\left(L^{-1}\vec{f}\right) \\ \vec{f} = \dfrac{1}{N_1}\sum_{\vec{X}^{(i)}\in D_1}\vec{X}^{(i)} \end{cases} \quad \begin{cases} \vec{v} = L^{-T}\left(L^{-1}\vec{g}\right) \\ \vec{g} = \dfrac{1}{N_0}\sum_{\vec{X}^{(i)}\in D_0}\vec{X}^{(i)} \end{cases} \tag{71}$$

Where, $\vec{f}$ and $\vec{g}$ represent the means of sample data in $D_1$ and $D_0$; $N_1$ and $N_0$ are sample numbers in $D_1$ and $D_0$; $\vec{\alpha}$ can be selected and calculated by Expression (63).

③ Compute the deviation vector $\vec{w}$ and the intermediate vector $\vec{c}$

$$\begin{cases} \vec{w} = L^{-T}\left(L^{-1}\vec{h}\right) \\ \vec{h} = \dfrac{1}{M_0}\sum_{\vec{X}^{(i)}\in S_0}\vec{X}^{(i)} - \dfrac{1}{M_1}\sum_{\vec{X}^{(i)}\in S_1}\vec{X}^{(i)} \end{cases} \quad \begin{cases} S_1 = \left\{\vec{X}^{(i)}\,\middle|\,\vec{X}^{(i)}\in D_1, \vec{\alpha}^T\vec{X}^{(i)} < \mu_1\right\}_{i=1}^{M_1} \\ S_0 = \left\{\vec{X}^{(i)}\,\middle|\,\vec{X}^{(i)}\in D_0, \vec{\alpha}^T\vec{X}^{(i)} > \mu_0\right\}_{i=1}^{M_0} \end{cases} \tag{72}$$

Where, $\vec{h}$ is the deviation of the sample data and only the sample data in $S_1$ and $S_0$ are calculated. $\mu_1 = \vec{\alpha}^T\vec{f}$ and $\mu_0 = \vec{\alpha}^T\vec{g}$, represent projections of the means of the sample data in $D_1$ and $D_0$ on $\vec{\alpha}$. After $\vec{u}$, $\vec{v}$, $\vec{w}$ are obtained, $\vec{c}$ can be obtained by Expression (64).

④ Calculate current values of the objective function and the constraint

Based on the optimization model (62), the current values of $J[\vec{p},\vec{q}]$ and $G[\vec{p},\vec{q}]$ can be calculated by

$$J_0 = \vec{c}^T K \vec{\alpha}, \qquad G_0 = \vec{u}^T K \vec{v} \tag{73}$$

(4) Calculate gradient vectors of the objective function and the constraint

Apply the Expressions (68, 69) to calculate $\partial J/\partial p_e$, $\partial J/\partial q_e$, $\partial G/\partial p_e$ and $\partial G/\partial q_e$. Then, express them as compact gradient vectors $\nabla_{\vec{p}}J$, $\nabla_{\vec{q}}J$, $\nabla_{\vec{p}}G$ and $\nabla_{\vec{q}}G$. Here, $\nabla_{\vec{p}}J$ is defined as $\nabla_{\vec{p}}J = [\partial J/\partial p_1 \quad \partial J/\partial p_2 \quad \cdots \quad \partial J/\partial p_M]^T$ and the other gradient vector definitions are similar. In the Expressions (68, 69), $K_p^{(e)}$ and $K_q^{(e)}$ are only determined by element interpolation basis functions and are constant matrices independent of design variables. So, $K_p^{(e)}$ and $K_q^{(e)}$ can be calculated prior to the optimization, and gradient vectors of $J[\vec{p},\vec{q}]$ and $G[\vec{p},\vec{q}]$ can be achieved through the mapping relationship between the local and global node numbers.

(5) Obtain increments of design variables by solving the sequential linearization optimization model

① Construct the sequential linearization optimization model

$$\min_{\vec{x}_p, \vec{x}_q} \quad J[\vec{x}_p, \vec{x}_q] = J_0 + \left(\nabla_{\vec{p}}J\right)^T \vec{x}_p + \left(\nabla_{\vec{q}}J\right)^T \vec{x}_q$$

s.t.

$$G[\vec{x}_p, \vec{x}_q] = G_0 + \left(\nabla_{\vec{p}}G\right)^T \vec{x}_p + \left(\nabla_{\vec{q}}G\right)^T \vec{x}_q \leq 0 \tag{74}$$

$$\sum_{i=1}^{Ne} x_{p,i} = Tolx_p, \qquad \sum_{e=1}^{Ne} x_{q,i} = Tolx_q$$

$$x_{p\min,i} \leq x_{p,i} \leq \Delta x_{\max}, \quad x_{q\min,i} \leq x_{q,i} \leq \Delta x_{\max} \qquad i = 1,2\cdots Ne$$

Where, design variables $\vec{p} \in R^M$, $\vec{q} \in R^M$; $\vec{x}_p$ and $\vec{x}_q$ are increments of the design variables, and their $i-th$ components are $x_{p,i}$ and $x_{q,i}$; $Tolx_p$, $Tolx_q$, and $x_{p\min,i}$, $x_{q\min,i}$ can be calculated by

$$\begin{cases} Tolx_p = Tolp - \sum_{i=1}^{Ne} p_i \\ Tolx_q = Tolq - \sum_{i=1}^{Ne} q_i \end{cases} \quad \begin{cases} x_{p\min,i} = \max(p_{\min} - p_i, -\Delta x_{\max}) \\ x_{q\min,i} = \max(q_{\min} - q_i, -\Delta x_{\max}) \end{cases}$$

② Solve the sequential linearization optimization model (74) to obtain $\vec{x}_p$ and $\vec{x}_q$

When solving Expression (74), slack variables are added to $G[\vec{x}_p, \vec{x}_q] \leq 0$ to facilitate the initial feasible solution construction.

(6) Determine whether to end the optimization iteration
   ① Store design variables, the objective function and the constraint function of the previous step of the sequential linearization optimization.
   Store design variables $\vec{p}^{old} = \vec{p}$, $\vec{q}^{old} = \vec{q}$, the objective function value $J_0^{old} = J_0$, and the constraint function value $G_0^{old} = G_0$.
   ② Update design variables and the objective function value.
   Let $\vec{p} = \vec{p} + \vec{x}_p$, $\vec{q} = \vec{q} + \vec{x}_q$, then execute step (3) to update the objective function value $J_0$.
   ③ Determine whether to end the iteration.
   If $|J_0 - J_0^{old}| \leq \varepsilon_J$ or $\max[\|\vec{x}_p\|_\infty, \|\vec{x}_p\|_\infty] \leq \varepsilon_x$, then end the iteration. Otherwise, if $J_0 < J_0^{old}$, go to the step (4) to continue the iteration; if $J_0 \geq J_0^{old}$, reduce the moving limits of design variables by letting $\Delta x_{\max} = \gamma \cdot \Delta x_{\max}$, here $\gamma = 0.5 \sim 0.85$, then go to step (5) to iteratively calculate design variable increments $\vec{x}_p$ and $\vec{x}_q$.

## 6. Algorithm Implementation and Image Classifier

Image classification is to determine if an image has certain given features and can be realized by algorithms for extracting the feature information of the image. Applying the mutual-energy inner product to extract image features has an advantage of enhancing feature information and suppressing other high-frequency noise. If we select multiple features of an image, we can design multiple mutual-energy inner products and each mutual-energy inner product can be regarded as one feature coordinate of the image. Using multiple mutual-energy inner products to characterize an image is equivalent to using multiple feature coordinates to describe the image, or equivalent to representing the high-dimensional image in a low-dimensional space, reducing the dimensionality of image data.

This part will discuss the implementation of the **Optimization Algorithm I** and its application in 2-D grayscale image classification. Assume that, each sample $X^{(i)}(\vec{x})$ in the training datasets $D_1$ and $D_0$ is a 2-D grayscale image; the domain $\Omega$ occupied by the image is rectangular; each image is expressed by $n_1 \times n_2$ pixels; and each pixel is a square with a side length 1. In this case, $\vec{x} \in \Omega \in R^2$ and $\Omega = \{(x_1, x_2) | 0 \leq x_1 \leq n_1, 0 \leq x_2 \leq n_2\}$.

### 6.1 Vectorized Implementation of Optimization Algorithm I

While using FEM to discretize the design domain, we regard each pixel as a finite element and divide the domain $\Omega$ into $n_1 \times n_2$ quadrilateral elements $\Omega_s^{(e)}$, i.e. $e = 1, 2 \cdots Ne$ and $Ne = n_1 \times n_2$. In $\Omega$, the global element numbering uses column priority, where the upper left corner element is numbered 1 and the lower right corner element is numbered $Ne$. A planar quadrilateral element is used to interpolate deformation functions. Each element has 4 nodes, so the total number of nodes is $M = (n_1 + 1) \times (n_2 + 1)$, and the total number of boundary nodes is $n_\Gamma = 2 \times (n_1 + n_2)$. The global node numbering also uses column priority, where the upper left corner node is numbered 1 and the lower right corner node is numbered $M$. The interpolation basis functions of the quadrilateral element are

$$N_i(\vec{\xi}) = \frac{1}{4}(1 + \xi_1^{(i)}\xi_1)(1 + \xi_2^{(i)}\xi_2) \qquad i = 1, 2, 3, 4 \qquad (75)$$

where the domain of definition is square and is expressed as $\Omega_\xi = \{(\xi_1, \xi_2) | -1 \leq \xi_1 \leq 1, -1 \leq \xi_2 \leq 1\}$. The element nodes are 4 corner points of the quadrilateral. The node with the coordinate $(-1,-1)$ is numbered 1, in counter-clockwise order, and other nodes with the coordinates $(+1,-1)$, $(+1,+1)$, $(-1,+1)$ are numbered 2, 3 and 4, respectively. The interpolation basis function $N_i(\vec{\xi})$ corresponds to the $i-th$ node, where $(\xi_1^{(i)}, \xi_2^{(i)})$ is the corresponding node coordinate. The mapping relationship between the element node numbers and the global node numbers can be described by an $Ne \times 4$ matrix $\Theta$, and its $i-th$ row corresponds to the $i-th$ element. If $\Theta_{i,j}$ denotes its entry at the $i-th$ row and the $j-th$ column, then $\Theta_{i,1}, \Theta_{i,2}, \Theta_{i,3}, \Theta_{i,4}$ are the global node numbers corresponding to the element node numbers 1, 2, 3 and 4 of the $i-th$ element. So, we have

$$\begin{cases} \Theta_{i,1} = i + 1 + r \\ \Theta_{i,2} = \Theta_{i,1} + n_1 + 1 \end{cases} \quad \begin{cases} \Theta_{i,3} = \Theta_{i,1} + n_1 \\ \Theta_{i,4} = \Theta_{i,1} - 1 \end{cases} \tag{76}$$

Where $r$ is a module when $i$ is divided by $n_1$. Since all elements are same squares, the isoparametric transformation $\vec{x}^{(e)}(\vec{\xi})$ in Expression (51) is actually a scaling transformation. Through substituting $N_i(\vec{\xi})$ into the Expressions (52, 55), we can find that the coefficient matrices $K_p^{(e)}$ and $K_q^{(e)}$ are independent of the element node numbers. So, we use $K_p$ and $K_q$ to express $K_p^{(e)}$ and $K_q^{(e)}$, and calculate them directly by

$$K_p = \frac{1}{24}\begin{bmatrix} 4 & -1 & -2 & -1 \\ -1 & 4 & -1 & -2 \\ -2 & -1 & 4 & -1 \\ -1 & -2 & -1 & 4 \end{bmatrix}, \quad K_q = \frac{1}{36}\begin{bmatrix} 4 & 2 & 1 & 2 \\ 2 & 4 & 2 & 1 \\ 1 & 2 & 4 & 2 \\ 2 & 1 & 2 & 4 \end{bmatrix} \tag{77}$$

When a side of an element overlaps with the boundary of the domain $\Omega$, the influence of the boundary conditions $l[u(\vec{x})] = 0$ in Expression (1) on $K_s^{(e)}$ should be considered, so a $4 \times 4$ matrix $K_\sigma^{(e)}$ should be calculated. Assume that the $j-th$ side of the element overlaps with the boundary of $\Omega$ and the entry in the $i-th$ row and $j-th$ column of $K_\sigma^{(e)}$ is $K_{\sigma i,j}^{(e)}$. Then, non-zero entries in $K_\sigma^{(e)}$ can be calculated by

$$K_{\sigma j,j}^{(e)} = K_{\sigma \hat{j},\hat{j}}^{(e)} = \frac{2}{3}\sigma_{j,\hat{j}}^{(e)}, \quad K_{\sigma j,\hat{j}}^{(e)} = K_{\sigma \hat{j},j}^{(e)} = \frac{1}{3}\sigma_{j,\hat{j}}^{(e)} \tag{78}$$

In Expression (78), the subscripts $j$ and $\hat{j}$ stand for the starting and end points of the $j-th$ side of the element, where the starting point is the element node numbered $j$ and the end point is determined along the side in counterclockwise order; $\sigma_{j,\hat{j}}^{(e)}$ is a constant, equal to the approximate value of $\sigma(\vec{x})$ on the $j-th$ side. In this paper, we handle the influence of $l[u(\vec{x})] = 0$ on $K_s^{(e)}$ during assembling the global stiffness matrix. We just simply replace the subscripts $j$ and $\hat{j}$ of $K_\sigma^{(e)}$ in Expression (78) with global node numbers, then directly use them to assemble the global stiffness matrix.

Because each element corresponds to a pixel, we can assume that its grayscale value is a constant $X_{gray}^{(i)(e)}$. In this way, a sample image $X^{(i)}(\vec{x})$ can be expressed as $\vec{X}_{gray}^{(i)} = [X_{gray}^{(i)(1)} \ X_{gray}^{(i)(2)} \cdots X_{gray}^{(i)(e)} \cdots X_{gray}^{(i)(Ne)}]^T$. Through substituting Expression (75) into Expression (70), the relationship between element node vectors and image grayscale values can be obtained

$$\vec{X}^{(i)(e)} = X_{gray}^{(i)(e)} \vec{F} \tag{79}$$

Where $\vec{F} = 0.25 \times [1\ 1\ 1\ 1]^T$, which can be regarded as mapping coefficients from the image grayscale to the element node vector.

While using element stiffness matrices $K_s^{(e)}$ and element node vectors $\vec{X}^{(i)(e)}$ to assemble the global stiffness matrix $K$ and the global node vector $\vec{X}^{(i)}$, the functions for generating a sparse matrix in MATLAB or the Python Scipy module can be used, and the input arguments include the row index vector, the column index vector and the values of non-zero entries. More importantly, these sparse matrix generation functions can sum non-zero entries with the same indexes, which is consistent with the process of assembling $K$ and $\vec{X}^{(i)}$.

In order to convert the image grayscale vector $\vec{X}_{gray}^{(i)}$ to the global node vector $\vec{X}^{(i)}$, a $4 \times Ne$ matrix $F\left(\vec{X}_{gray}^{(i)}\right)^{\mathrm{T}}$ should first be calculated, whose $e-th$ column corresponds to the element node vector $\vec{X}^{(i)(e)}$. Then, $F\left(\vec{X}_{gray}^{(i)}\right)^{\mathrm{T}}$ is converted to a $4Ne$-dimensional column vector $\vec{V}_X$ in column-major order. Obviously, if we divide components of $\vec{V}_X$ into multiple groups in sequence and each group includes 4 components, then the $e-th$ group corresponds to the element node vector $\vec{X}^{(i)(e)}$. $\vec{V}_X$ can be calculated by

$$\vec{V}_X = reshape\left(F\left(\vec{X}_{gray}^{(i)}\right)^{\mathrm{T}}, 4Ne, 1\right) \tag{80}$$

Where the function $reshape(A, m, n)$ can convert the dimension of the matrix $A$ into $m \times n$ while remaining the total number of entries unchanged.

Through the mapping matrix $\Theta$, position indexes of components of $\vec{V}_X$ in the global node vector $\vec{X}^{(i)}$ can be obtained. We transpose the $Ne \times 4$ matrix $\Theta$ to the $4 \times Ne$ matrix $\Theta^T$, whose $e-th$ column corresponds to global node numbers of the $e-th$ element, and then convert $\Theta^T$ to a $4Ne$-dimensional column vector $\vec{I}_X$ in column-major order. $\vec{I}_X$ can be figured out by

$$\vec{I}_X = reshape\left(\Theta^T, 4Ne, 1\right) \tag{81}$$

$\vec{I}_X$ is the row index vector to generate $\vec{X}^{(i)}$ by a sparse matrix generation function. Since $\vec{X}^{(i)}$ has only one column, we use $\vec{J}_X$ to denote a $4Ne$-dimensional column index vector and set all the components of $\vec{J}_X$ to 1. Through substituting $\vec{V}_X$, $\vec{I}_X$, $\vec{J}_X$ into the sparse matrix generation function, we can yield $\vec{X}^{(i)}$.

Similarly, the global stiffness matrix $K$ can be assembled by using the sparse matrix generation function. A vector $\vec{V}_K^{(pq)}$ related to $K_s^{(e)}$ should be first calculated by

$$\vec{V}_K^{(pq)} = reshape\left(K_s^{(pq)}, 16Ne, 1\right), \quad K_s^{(pq)} = K_p \otimes \vec{p}^T + K_q \otimes \vec{q}^T \tag{82}$$

Where, the operator $\otimes$ denotes the Kronecker product of matrices; $K_s^{(pq)}$ is a $4 \times 4Ne$ matrix; $\vec{p}$ and $\vec{q}$ are design variables. If $K_s^{(pq)}$ is divided into multiple blocks from left to right and each block is a $4 \times 4$ matrix, the $e-th$ block is the calculation result of the first two terms of $K_s^{(e)}$ in Expression (54), without including $K_\sigma^{(e)}$. Therefore, if $\vec{V}_K^{(pq)}$ are divided into multiple blocks in sequence and each block includes 16 components, the $e-th$ block will correspond to a 1-dimensional vector converted from the $e-th$ element stiffness matrix in column priority. We set $\vec{1} = \begin{bmatrix} 1 & 1 & 1 & 1 \end{bmatrix}^T$, and use $\vec{I}_K^{(pq)}$, $\vec{J}_K^{(pq)}$ to denote row indexes and column indexes of entries in the global stiffness matrix. Then, $\vec{I}_K^{(pq)}$, $\vec{J}_K^{(pq)}$ corresponding to components of $\vec{V}_K^{(pq)}$ can be calculated by

$$\vec{I}_K^{(pq)} = reshape\left(\Theta^T \otimes \vec{1}^T, 16Ne, 1\right), \quad \vec{J}_K^{(pq)} = reshape\left(\Theta^T \otimes \vec{1}, 16Ne, 1\right) \tag{83}$$

As mentioned above, the constraint on the design boundary $\Gamma$ can generate additional stiffness $K_\sigma^{(e)}$ for adjacent elements. If we regard an element side overlapping with $\Gamma$ as a 2-node line element, then its stiffness matrix will be a $2 \times 2$ matrix $K_\Gamma^{(e)}$, which can be figured out by Expression (78). Similarly, these line element stiffness matrices can be assembled into the global stiffness matrix. While designing an image classifier based on the mutual-energy inner products, we set a fixed boundary for Expression (1), i.e. $u(\vec{x}) = 0$ $(\vec{x} \in \Gamma)$. This boundary condition can be handled by adding a relatively large number $\sigma_0$ to the diagonal entries of $K$, where its diagonal entries correspond to boundary node numbers. The sparse matrix generation function is used to implement this boundary condition. First, we set the dimension of the vector $\vec{V}_K^{(\sigma_0)}$ as $n_\Gamma$, which is the total number of boundary nodes, and set all components of $\vec{V}_K^{(\sigma_0)}$ to $\sigma_0$. Meanwhile, we let the $n_\Gamma$-dimensional row and column index vectors be the same, i.e. $\vec{I}_K^{(\sigma_0)} = \vec{J}_K^{(\sigma_0)}$, and set their components to be the boundary node numbers. Finally, we combine $\vec{V}_K^{(pq)}$ and $\vec{V}_K^{(\sigma_0)}$, $\vec{I}_K^{(pq)}$ and $\vec{I}_K^{(\sigma_0)}$, $\vec{J}_K^{(pq)}$ and $\vec{J}_K^{(\sigma_0)}$, respectively, and input them to the sparse matrix generation function to obtain $K$.

Based on the Expressions (68, 69), gradients of the objective and the constraint can be efficiently obtained by using $\Theta$. For example, if we have two $M$-dimensional global node vectors $\vec{c}$ and $\vec{\alpha}$, and adopt fancy indexing to generate two $Ne \times 4$ matrices $N_c = \vec{c}(\Theta)$ and $N_\alpha = \vec{\alpha}(\Theta)$, whose $e-th$ rows correspond to node vectors of the $e-th$ element. According to Expression (69), the objective function gradients $\nabla_{\vec{p}} J$, $\nabla_{\vec{q}} J$ can be calculated by

$$\nabla_{\vec{p}} J = -sum(N_c K_p \cdot * N_\alpha, 2), \quad \nabla_{\vec{q}} J = -sum(N_c K_q \cdot * N_\alpha, 2) \tag{84}$$

Where, $\cdot *$ stands for multiplying corresponding entries of matrices, and $sum(A, 2)$ is summing the rows of a matrix to obtain a column vector. Mathematically, Expression (84) can be written as $\nabla_{\vec{p}} J = diag(N_c K_p N_\alpha^T)$ and $\nabla_{\vec{p}} J = diag(N_c K_q N_\alpha^T)$, where the function $diag(\ )$ is to extract the main diagonal entries from a square matrix. Similarly, constraint function gradients $\nabla_{\vec{p}} G$, $\nabla_{\vec{q}} G$ can be calculated by replacing $\vec{c}$, $\vec{\alpha}$ with $\vec{u}$, $\vec{v}$.

## 6.2 Image classifier

For a given training dataset $D$, in order to use the **Optimization Algorithm I** to construct the mutual-energy inner product coordinate axes $\vec{\alpha}_m (m = 1, 2 \cdots N_\alpha)$, we select the subset $D_s$ of $D$ as the reference training set and select the mean of samples of the class "0" or the class "1" in $D_s$ or a combination of these means as the reference feature $\vec{\alpha}$. The subset $D_s$ is gradually generated as the coordinate $\vec{\alpha}_m$ is generated. Prior to generating the coordinate $\vec{\alpha}_{m+1}$, $m$ mutual-energy inner product coordinate axes $\vec{\alpha}_1, \vec{\alpha}_2 \cdots \vec{\alpha}_m$ have been generated and there are $m$ subsets $D_s^{(1)}, D_s^{(2)} \cdots D_s^{(m)}$. One of the $m$ subsets is selected as a subset $D_s$ to generate the coordinate $\vec{\alpha}_{m+1}$. In order to explain how the generation of new axes work, we use a set $S_T$ to manage the $m$ generated subsets, i.e. $S_T = \{D_s^{(1)}, D_s^{(2)} \cdots D_s^{(m)}\}_1^m$. If $D_s^{(i)}$ has $M_0^{(i)}$ samples of the class "0" and $M_1^{(i)}$ samples of the class "1", the subset $D_s^{(k)}$ in $S_T$ is taken as the reference training sample set $D_s$ to generate $\vec{\alpha}_{m+1}$ and its index $k$ satisfies

$$k = \underset{i}{\operatorname{argmax}} \ \min(M_0^{(i)}, M_1^{(i)}) \tag{85}$$

After determining the subset $D_s$ and the reference feature $\vec{\alpha}$, $\vec{\alpha}_{m+1}$ can be obtained by the **Optimization Algorithm I**. Next, we divide $D_s$ into two subsets. First, for each sample $\vec{X}^{(i)}$ in $D_s$, calculate its coordinate component $z_{m+1}^{(i)}$ on the axis $\vec{\alpha}_{m+1}$ by $z_{m+1}^{(i)} = \langle \vec{\alpha}_{m+1}, \vec{X}^{(i)} \rangle$; calculate $\mu_{0,m+1}$ and $\mu_{1,m+1}$, the means of samples of the class "0" and the class "1"; and set a threshold $z_{m+1}^{th} = (\mu_{0,m+1} + \mu_{1,m+1})/2$. Second, according to $z_{m+1}^{(i)}$ and $z_{m+1}^{th}$, divide $D_s$ into two subsets satisfying $D_{sI} = \{\vec{X}^{(i)} | \vec{X}^{(i)} \in D_s, \ z_{m+1}^{(i)} \leq z_{m+1}^{th}\}$ and $D_{sII} = \{\vec{X}^{(i)} | \vec{X}^{(i)} \in D_s, \ z_{m+1}^{(i)} > z_{m+1}^{th}\}$. Finally, add $D_{sI}$ and $D_{sII}$ into $S_T$, and delete $D_s^{(k)}$ from $S_T$. At this time, $S_T$ contains $m+1$ training sample subsets, and one of them will be selected to calculate the coordinate axis $\vec{\alpha}_{m+2}$.

The following summarizes detailed steps of generating mutual-energy inner product feature coordinates.

**Algorithm II: Mutual-energy inner product feature coordinates generation**

(1) Let $m = 0$ and $S_T = \{D\}$;

(2) According to Expression (85), select $D_s$ in $S_T$ to generate the coordinate axis $\vec{\alpha}_{m+1}$ and delete $D_s^{(k)}$ from $S_T$;

(3) Adopt the **Optimization Algorithm I** to calculate $\vec{\alpha}_{m+1}$ based on the determined reference subset $D_s$ and the selected reference feature $\vec{\alpha}$;

(4) For each sample $\vec{X}^{(i)}$ in $D_s$, calculate its coordinate components $z_{m+1}^{(i)}$ on the axis $\vec{\alpha}_{m+1}$, the means $\mu_{0,m+1}$ and $\mu_{1,m+1}$ of the class "0" and the class "1", as well as the threshold $z_{m+1}^{th} = (\mu_{0,m+1} + \mu_{1,m+1})/2$;

(5) According to $z_{m+1}^{(i)}$ and $z_{m+1}^{th}$, divide $D_s$ into two subsets $D_{sI}$ and $D_{sII}$, and add them into $S_T$;

(6) Judge: if $m < N_\alpha$, set $m = m + 1$ and go to the Step (2); otherwise, end.

After generating mutual-energy inner product coordinate axes $\vec{\alpha}_m (m = 1, 2 \cdots N_\alpha)$ by **Algorithm II**, the coordinate components $z_m^{(i)} (m = 1, 2 \cdots N_m)$ of each sample in $D$ can be calculated and is represented by a feature vector $\vec{z}^{(i)} = \begin{bmatrix} z_1^{(i)} & z_2^{(i)} & \cdots & z_{N_m}^{(i)} \end{bmatrix}^T$. Based on $\vec{z}^{(i)}$, a simple Gaussian classifier is used to classify images. We use $D_j$ to represent a training dataset comprising $M_j$ samples, where the subscript $j$ is the class index of samples. Gaussian classifier can be used to classify samples into multiple classes. We use $y^{(i)} = j (j = 0, 1 \cdots C - 1)$ to indicate the class of a sample and use $C$ to denote a total number of classes. In $D$, the probability of the class $y^{(i)} = j$ is

$$p(y^{(i)} = j) = \frac{M_j}{\sum_{k=0}^{C-1} M_k} \tag{86}$$

Furthermore, it is assumed that, for samples in the same class, their feature vectors $\vec{z}^{(i)}$ follow the Gaussian distribution

$$p(\vec{z}^{(i)} | y^{(i)} = j) \sim N(\vec{z}^{(i)} | \vec{\mu}_j, \Sigma_j) \tag{87}$$

Where, $\vec{\mu}_j$ is the mean of $\vec{z}^{(i)}$; $\Sigma_j$ is the covariance matrix of $\vec{z}^{(i)}$; and the subscript $j$ corresponds to the class $y^{(i)} = j$. Using the training sample dataset $D_j$, their maximum likelihood estimates can be calculated by [38]

$$\vec{\mu}_j = \frac{1}{M_j} \sum_{i=1}^{M_j} \vec{z}^{(i)}, \quad \Sigma_j = \frac{1}{M_j} \sum_{i=1}^{M_j} (\vec{z}^{(i)} - \vec{\mu}_j)(\vec{z}^{(i)} - \vec{\mu}_j)^T \tag{88}$$

Here, $z^{(i)} \in D_j$. Based on the Expressions (86, 87), when giving the feature vector of a sample, the posterior probability of the sample belonging to the class $y = j$ is

$$p(y = j | \vec{z}) = \frac{e^{\beta_j(\vec{z})}}{\sum_{i=1}^{C} e^{\beta_i(\vec{z})}} \tag{89}$$

Where, $p(y = j | \vec{z})$ is the posterior probability, and $\beta_j(\vec{z})$ can be expressed as

$$\begin{cases} \beta_j(\vec{z}) = \dfrac{1}{2}\vec{z}^T H_j \vec{z} + \vec{b}_j^T \vec{z} + c_j \\ H_j = -\Sigma_j^{-1} \quad \vec{b}_j = \Sigma_j^{-1}\vec{\mu}_j \\ c_j = -\dfrac{1}{2}\vec{\mu}_j^T \Sigma_j^{-1}\vec{\mu}_j - \dfrac{1}{2}\ln\left(|\Sigma_j|\right) + \ln\left(p(y=j)\right) \end{cases} \quad (90)$$

Finally, the class of the sample is determined based on the posterior probability

$$y = \arg\max_j p(y=j|\vec{z}) \quad j \in \{0,1\cdots C-1\} \quad (91)$$

## 6.3 Numerical examples

The MNIST dataset has become one of benchmark datasets in machine learning. It comprises 60,000 sample images in the training set and 10,000 sample images in the test set, and each one is a 28-by-28-pixel grayscale image of handwritten digits 0-9. In this section, we will use the MNIST to design Gaussian image classifiers based on the **Optimization Algorithm I**.

Before designing Gaussian image classifiers, image preprocessing is conducted to align the image centroids and normalize the sample images. In the **Optimization Algorithm I**, selected parameters are $\lambda = 0.3$, $p_{\min} = q_{\min} = 10^{-3}$, $Tolp = Tolq = 2.0$, $\sigma_0 = 10^5$, $\Delta x_{\max} = 0.08$, $\varepsilon_x = 8\times 10^{-4}$, and $\varepsilon_J = 10^{-7}$.

### 6.3.1 Binary Gaussian classifier: identify digits "0" and "1"

The MINST training set comprises 6,742 samples "1" and 5,923 samples "0". We select the difference between the means of samples "1" and "0" as the reference feature, i.e. $\vec{\alpha} = \vec{u} - \vec{v}$. The **Optimization Algorithm I** converges after 166 iterations. The means of samples "1" and "0", design variables, and the reference feature coordinate $\alpha(\vec{x})$, are visualized in Figures 1-3. Due to obvious difference in the mean feature, digits "0" and "1" can be identified using only one mutual-energy inner product coordinate $\vec{\alpha}$. The histogram 4(a) shows the training sample distribution in accordance with the components on $\vec{\alpha}$. Figure 5(a) gives the Confusion Matrix of classification results, where the horizontal and vertical axes correspond to the target class and the output class of the classifier, respectively. In the Confusion Matrix, the column on the far right shows the precision of all the examples predicted to belong to each class, and the row at the bottom shows the recall of all the examples belonging to each class; the entry in the bottom right shows the overall accuracy; diagonal entries are correctly classified numbers of digits "0" and "1" and off-diagonal entries correspond to the wrong classifications. This binary Gaussian classifier on the training set achieves a very high overall accuracy of 99.66%, shown at the bottom right of the Confusion Matrix.

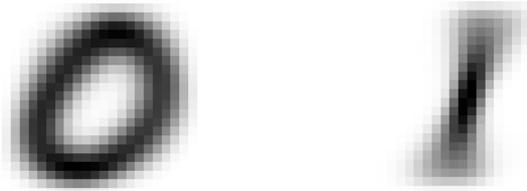

Figure 1 The means of samples "0" and "1"

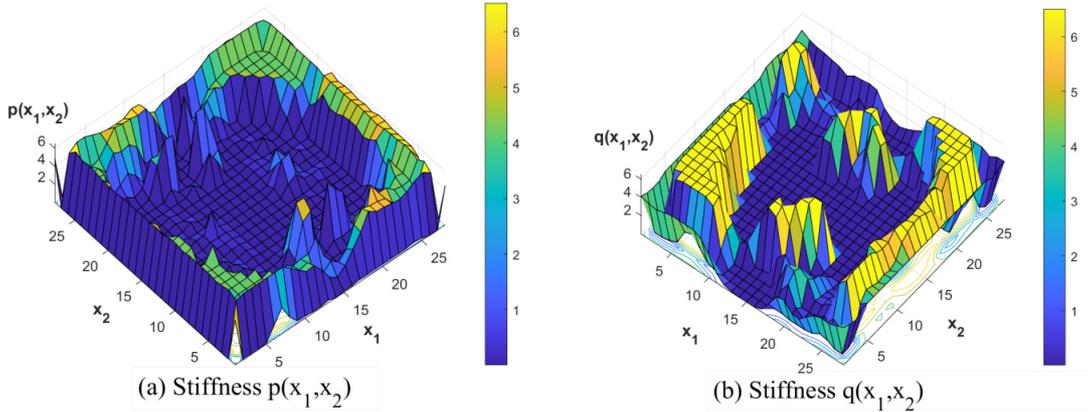

Figure 2 Design variables

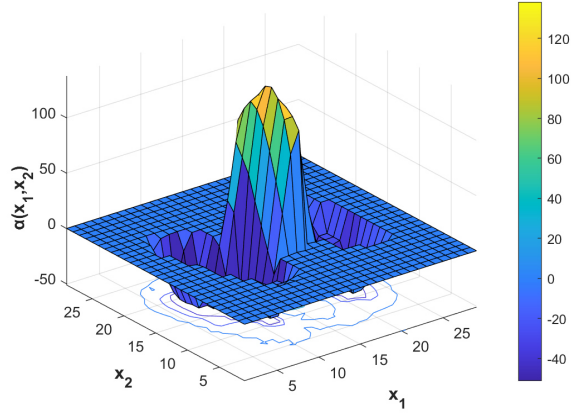

Figure 3 Reference feature coordinate axis $\alpha(\vec{x})$

The binary Gaussian classifier is tested on the MINST test set, which comprises 1,135 samples "1" and 980 samples "0". The test results are visualized in Figures 4(b) and 5(b). Its overall accuracy can reach 99.91%, higher than that on the training set.

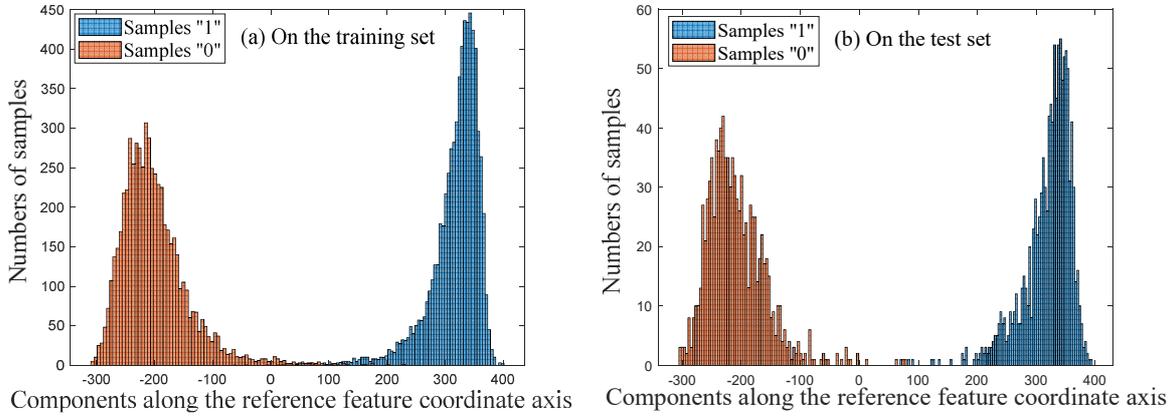

Figure 4 Distribution of samples "1" and "0"

Figure 5 Confusion Matrix

**6.3.2 Binary Gaussian classifier: identify digits "0" and "2"**

The MINST training set comprises 5,958 samples "2" and 5,923 samples "0", and the MINST test set comprises 1,032 samples "2" and 980 samples "0". Similar to the previous classifier, the reference feature is also selected as $\vec{\alpha} = \vec{u} - \vec{v}$. The difference in mean features of digits "2" and "0" is not as significant as that of digits "1" and "0". If only one mutual-energy inner product coordinate is used for classification, the accuracy is only 96.72% on the training set and 97.81% on the test set. In order to improve the classification accuracy, we use **Algorithm II** to generate 60 mutual-energy inner product coordinates based on the training sample set and its subsets, and construct a 60-dimensional Gaussian classifier. Confusion Matrices of classification results are given in Figure 6 (a, b), showing an overall accuracy of 99.55% on the training set and a higher overall accuracy of 99.85% on the test set.

**6.3.3 Binary Gaussian classifier: identify digits "3" and "4"**

The MINST training set comprises 6,131 samples "3" and 5,842 samples "4", and the MINST test set comprises 1,010 samples "3" and 982 samples "4". Here, we select the means of samples "3" and "4" as reference features, i.e. $\vec{\alpha} = \vec{u}$ and $\vec{\alpha} = \vec{v}$, and then use **Algorithm II** to generate 50 mutual-energy inner product coordinates, respectively, finally forming 100 classification coordinates. Because these coordinates are not linearly independent, we use matrix singular value decomposition to construct a 50-dimensional Gaussian classifier. Figure 6(c, d) is Confusion Matrices, showing an overall accuracy of 99.67% on the training set and a higher overall accuracy of 99.80% on the test set.

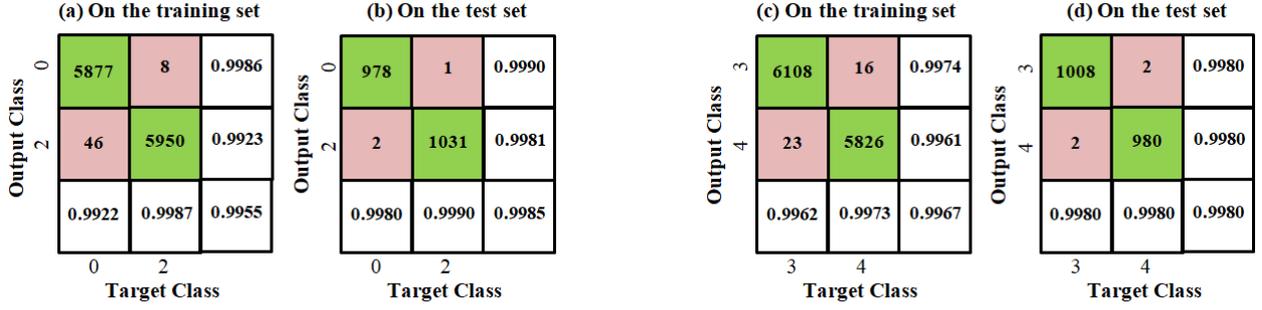

Figure 6 Confusion Matrix

### 6.3.4 Multiclass Gaussian classifier: identify digits "0", "1", "2", "3" and "4"

In the training set, we select one digit from samples "0", "1", "2", "3" and "4" as the first class and the other training samples of the five digits as the second class, and we take the two classes as the training sample set. Then, we select the difference between the means of samples in two classes as the reference feature, i.e. $\vec{\alpha} = \vec{u} - \vec{v}$, and use **Algorithm II** to generate 120 mutual-energy inner product coordinates. In this way, we construct 5 training sample sets and finally generate 600 coordinates. However, many of them are linearly dependent, in order to identify digits "0", "1", "2", "3" and "4", we use matrix singular value decomposition to reduce its dimensions from 600 to 60 and construct a 60-dimensional multiclass Gaussian classifier. Figure 7 shows an overall accuracy of 98.22% on the training set and a higher overall accuracy of 98.83% on the test set.

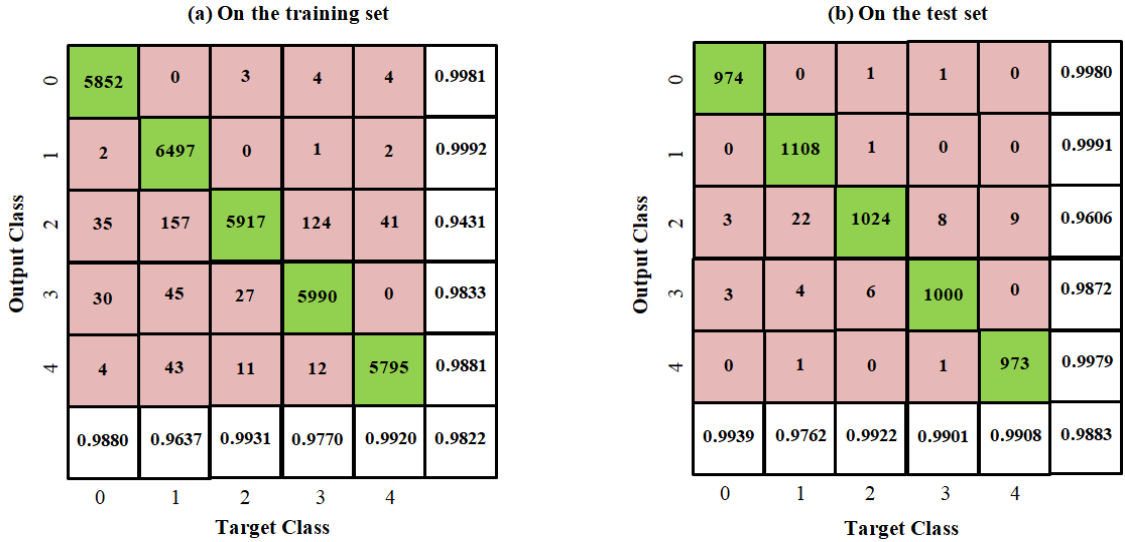

Figure 7 Confusion Matrix

## 7. Discussion

Based on the solution space of partial differential equations describing the vibration of a non-uniform membrane, the concept of mutual-energy inner product is defined. By expending the mutual-energy inner product as a superposition of eigenfunctions of the partial differential equations, an important property is found that the mutual-energy inner product has a significant advantage of enhancing low-frequency eigenfunction components and suppressing high-frequency eigenfunction components, compared with the Euclidean inner product.

In data classification, if the reference data features of samples belong to a low-frequency subspace of the set of the eigenfunctions, these data features can be extracted through the mutual-energy inner product, which can not only enhance feature information but also filter out high-frequency data noise. As a result, a mutual-energy inner product optimization model is built to extract feature coordinates of samples, which can enhance data features, reduce sample deviations, and regularize design variables. We make use of the minimum energy principle to eliminate constraints of partial differential equations in the optimization model and obtain an unconstrained optimization objective function. The objective function is a quadratic functional, which is convex with respect to variables that minimize the objective function, is concave with respect to variables that maximize the objective function, and is linear with respect to design variables. These properties facilitate the design of optimization algorithms.

FEM is used to discrete the design domain, and design variables of each element are set as constants. Based on these finite elements, gradients of the mutual-energy inner product relative to element design variables are analyzed, and a sequential linearization algorithm is constructed to solve the mutual-energy inner product optimization model. Algorithm implementation only involves solving equations including positive definite symmetric matrix when calculating intermediate variables and only needs to handle a few constraints in the nested linear optimization module, guaranteeing the stability and effectiveness of the algorithm.

The mutual-energy inner product optimization model is applied to extract feature coordinates of sample images and construct a low-dimensional coordinate system to represent sample images. Multiclass Gaussian classifiers are trained and tested to classify 2-D images. Here, only the means of the training sample set and its subsets are selected as reference features in the **Optimization Algorithm I**, and the vectorized implementation of the **Optimization Algorithm I** is discussed. Generating mutual-energy inner product coordinates by the optimization model and training or testing Gaussian classifiers are two independent steps. In training or testing Gaussian classifiers, calculating mutual-energy inner products can be converted into calculating the Euclidean inner products between the reference feature coordinates and sample data, not adding computational complexity to the Gaussian classifiers.

On the MINST dataset, the mutual-energy inner product feature coordinate extraction method is used to train a 1-dimensional two-class Gaussian classifier, a 50-dimensional two-class Gaussian classifier, a 60-dimensional two-class Gaussian classifier, and a 60-dimensional five-class Gaussian classifier, and good prediction results are achieved. The feature coordinate extraction method achieves a higher overall accuracy on the test set than that on the training set, indicating that the classification model is experiencing underfitting. This shows large potential in achievable accuracy of this method that has not yet been explored. In the future, convolution operation can be adopted to construct other image features, such as image edge features, local features, and multi-scale features, and these image features can be combined to generate mutual-energy inner product feature coordinate system. In addition, other ensemble classifiers, such as Bagging, AdaBoost, can be introduced to improve performances of image classifiers. Meanwhile, the feasibility of applying the mutual-energy inner product optimization method to the neural network will also be explored.

**Data Availability Statement:** Data and code are available upon request from the author.
**Conflicts of Interest:** The author declares no conflict of interest
**Acknowledgments:** Appreciate Professor Brian Barsky and Professor Stuart Russell for their encouragement during my difficult times.